\pdfoutput=1
\documentclass[11pt,a4paper]{article}
\usepackage[hyperref]{naaclhlt2019}
\usepackage{latexsym}

\aclfinalcopy 
\def\aclpaperid{1200} 

\usepackage{amsmath,amssymb}
\usepackage[T1]{fontenc}
\usepackage[utf8]{inputenc}
\usepackage{lmodern}
\usepackage{mathptmx}

\usepackage{soul}
\usepackage{latexsym}
\usepackage{todonotes}
\usepackage{booktabs}
\usepackage{enumitem}
\usepackage{pgfplots}
\usepackage{pgfplotstable}
\usepackage{subfig}

\usepackage{microtype}

\newcommand{\adlof}{{\textsf{VIPER}}}

\newcommand{\Removed}[1]{}
\definecolor{ao}{rgb}{0.0, 0.5, 0.0}

\title{Text Processing Like Humans Do:\\ Visually Attacking and Shielding NLP Systems}

\author{Steffen Eger\textsuperscript{$\dagger\ddagger$},~
G\"ozde G\"ul Şahin\textsuperscript{$\dagger\ddagger$},~
Andreas R\"uckl\'e\textsuperscript{$\dagger$},~
Ji-Ung Lee\textsuperscript{$\dagger\ddagger$},~
Claudia Schulz\textsuperscript{$\dagger\ddagger$},
\\
\textbf{Mohsen Mesgar\textsuperscript{$\dagger\ddagger$},}~
\textbf{Krishnkant Swarnkar\textsuperscript{$\dagger$},}~
\textbf{Edwin Simpson\textsuperscript{$\dagger$},}~
\textbf{Iryna Gurevych\textsuperscript{$\dagger\ddagger$}}\\[.3em]
	\textsuperscript{$\dagger$}Ubiquitous Knowledge Processing Lab (UKP-TUDA)\\
	\textsuperscript{$\ddagger$}Research Training Group AIPHES\\
	Department of Computer Science, Technische Universit\"{a}t Darmstadt\\
	\textsuperscript{$\dagger$}{\url{www.ukp.tu-darmstadt.de}}\\
	\textsuperscript{$\ddagger$}{\url{www.aiphes.tu-darmstadt.de}}\\
}

\begin{document}

\maketitle

\begin{abstract}
Visual modifications to text are often used to obfuscate offensive comments in social media (e.g., ``!d10t'') or 
as a writing style (``1337'' in ``leet speak''),
among other scenarios.
We consider this as a new type of adversarial attack in NLP,  a 
setting
to which humans are very robust, 
as our experiments with both simple and more difficult visual 
perturbations demonstrate.
We 
investigate 
the impact of visual adversarial attacks
on current NLP systems on \mbox{character-,} \mbox{word-,} and sentence-level tasks,
showing
that both neural and non-neural models are, in contrast to humans, 
extremely sensitive to such attacks, 
suffering performance decreases of up to 82\%. 
We then explore three shielding methods---visual character embeddings, adversarial training, and rule-based recovery---which
substantially improve the robustness of the models.
However, the shielding methods still fall behind performances achieved in 
non-attack scenarios, which demonstrates the difficulty of dealing with visual attacks.
\end{abstract}

\section{Introduction}
For humans, visual similarity can play a decisive role for assessing the meaning of characters. 
Evidence for this includes: 
the frequent swapping of similar looking characters in Internet slang or 
abusive comments,
creative trademark logos, 
and attack scenarios such as domain name spoofing
(see examples in Table~\ref{tbl:examples-intro}).

Recently, some NLP systems have exploited visual features to capture visual relationships among characters in compositional writing systems such as Chinese or Korean \cite{LiuLLN17}.
However, in more general cases, current neural NLP systems have no \mbox{built-in} notion of visual character similarity. 
Rather, they either treat characters as discrete units 
forming a word 
or 
they represent characters by randomly initialized embeddings and update them during training---typically in order to generate a character-based word representation that is robust to morphological variation or spelling mistakes \cite{Ma:2016}. 
Intriguingly, this marked distinction between human and machine processing can be exploited as a blind spot of NLP systems. 
For example, spammers might send malicious 
emails or post toxic comments to online discussion forums \cite{hosseini2017deceiving} by visually `perturbing' the input text in such a way that it is still easily recoverable by humans. 

\begin{table}
\centering
\includegraphics[scale=0.8]{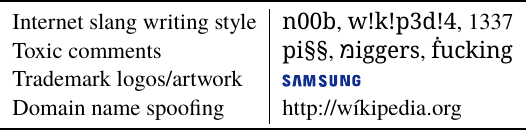}
\caption{Examples of text in which characters have been changed to visually similar ones.}
\label{tbl:examples-intro}
\end{table}

The issue 
of exposing and addressing the weaknesses of deep learning models to \emph{adversarial inputs}, i.e., perturbed versions of original input samples, has recently received considerable attention.  
For instance,  \newcite{Goodfellow:2015} showed that small perturbations in the pixels of an image can mislead a neural classifier to predict an incorrect label for the image. 
In NLP, \newcite{jia2018} 
inserted grammatically correct but semantically irrelevant paragraphs to stories to fool neural reading comprehension models. 
\newcite{SinghGR18} showed significant drops in the performance of neural models for question answering when using simple paraphrases of the original questions. 

Unlike previous NLP attack scenarios, visual attacks, 
i.e., the exchange of characters in the input with visually similar alternatives, 
have the following `advantages':
1) They do not require \emph{any} linguistic knowledge beyond the character level, making the attacks straightforwardly applicable across languages, domains, and tasks.
2) They are allegedly less damaging to human perception and understanding than, e.g., syntax errors or the insertion of negations \cite{hosseini2017deceiving}. 
3) They do not require knowledge of the attacked model's parameters or loss function \cite{hotFlip18}. 

In this work, we investigate to what extent recent state-of-the-art (SOTA) deep learning models are sensitive to  {visual} attacks and explore various shielding techniques. Our contributions are: 

\begin{itemize}[noitemsep]
  \item
  We introduce \adlof{}, a 
  Visual Perturber that
  randomly replaces characters in the input with their visual nearest neighbors in a visual embedding space. 
  \item
  We show that the performance of SOTA deep learning models substantially {drops} for various NLP tasks when attacked by \adlof{}. 
  On individual tasks (e.g., Chunking) and attack scenarios, our observed drops are up to 82\%.
  
  \item 
  We show that, in contrast to NLP systems,
  humans are only 
  mildly or not at all affected by visual perturbations. 
  \item
  We explore three methods to shield
  from visual attacks, \textit{viz.}, visual character embeddings, adversarial training \cite{Goodfellow:2015}, and rule-based recovery. We quantify to which degree and in which circumstances these are helpful. 
\end{itemize}
We point out that integrating visual knowledge with deep learning systems, as 
our visual character embeddings do, 
aims to make NLP models behave more like humans by  
taking cues directly from sensory information such as vision.\footnote{Code and data available from \url{https://github.com/UKPLab/naacl2019-like-humans-visual-attacks}}

\section{Related Work}
Our work connects to two strands of literature: adversarial attacks and visually informed character embeddings.
\paragraph{Adversarial Attacks}
are modifications to a classifier's input, that are designed to fool the system into making an incorrect decision, while 
the original meaning 
is still understood by a human observer.
Different forms of attacks have been studied in NLP and computer vision (CV), including at a character, syntactic, semantic and, in CV, the visual level. 
\citet{hotFlip18} propose a character flipping algorithm to generate adversarial examples and use it to trick a character-level neural classifier. They show that the accuracy  decreases significantly after a few manipulations if certain characters are swapped. 
Their character flipping approach requires very strong knowledge in the form of the attacked networks' gradients in a so-called white box 
attack setup. 
~\citet{chen2018attacking} 
find that reading comprehension systems often ignore important question terms, 
thus giving incorrect answers when these terms are replaced.  
~\citet{Belinkov:2018} show that neural machine translation systems 
break
for all kinds of noise to which humans are robust, such as reordering characters in words, keyboard typos and spelling mistakes.~\citet{Alzantot:2018} replace words by synonyms to fool text classifiers. \citet{Iyyer:2018} reorder sentences syntactically to generate adversarial examples. 

In contrast 
to those related works which perform attacks on the character level, our attacks 
allow 
perturbation of any character in a word while potentially minimizing impairment for humans. 
For example, the strongest attack in \citet{Belinkov:2018} is random shuffling of all characters, which is much more difficult to restore for humans. 

To cope with adversarial attacks, 
\emph{adversarial training} \cite{Goodfellow:2015} has been proposed as a standard remedy in which training data is augmented with data that is similar to the data used to attack the neural classifiers. 
\citet{rodriguez2018shielding} propose simple rule-based corrections to address a limited number of attacks, including obfuscation (e.g., ``idiots'' to ``!d10ts'') and negation (e.g., ``idiots'' to ``NOT idiots''). 
 Most other approaches have been explored in the context of CV, such as adding a stability objective during training \cite{Zheng2016} and distillation \cite{papernot2016}. However, methods to increase the robustness in CV have been shown to be less effective against more sophisticated attacks \cite{Carlini2017}. 

\paragraph{Visual Character Embeddings}
were originally proposed to address large character vocabularies 
in `compositional' languages like Chinese and Japanese. 
\citet{ShimadaKI16} and ~\citet{DaiC17} employ a convolutional autoencoder to generate image-based character embeddings (ICE) for Japanese and Chinese text and show improvement on author and publisher identification tasks. 
Similarly, ~\citet{LiuLLN17} 
create ICEs from a CNN and show that ICEs carry more semantic content and 
are more suitable for rare characters.
However, 
existing work on visual character embeddings 
has not used 
visual information to attack NLP systems or to 
them.

\section{Approach}
To investigate the effects of visual attacks
and propose methods for shielding, 
we introduce 
1) a visual text perturber,
2) three character embedding spaces, 
and 3) methods for obtaining word embeddings from character embeddings,  
used as input representations in some of our experiments. 

\subsection{Text perturbations}
Our visual perturber \adlof{} disturbs an input text in such a way that (ideally) it is still readable by humans but causes NLP systems to fail blatantly. We parametrize \adlof{} 
by a probability $p$ and a character embedding space, CES:\footnote{CES may be any `embedding space' that can be used to identify the nearest neighbors of characters.} 
For each character $c$ in the input text 
a flip decision is made (i.i.d.\ Bernoulli distributed with probability $p$), and if a replacement takes place, 
one of up to 20 nearest neighbors in the CES is chosen.\footnote{The probability of choosing one of the 20 neighbors of $c$ is proportional to its distance to $c$.}
Thus, we denote \adlof{} as taking two arguments:
\begin{align*}
  \text{\adlof{}} = \text{\adlof{}}(p,\text{CES}).
\end{align*}
{Note that \adlof{} is a black-box attacker as it does not require any knowledge of the attacked system.}
It would also be possible to design a 
more intelligent
perturber that only disturbs content words (or ``hot'' words), similar to~\citet{hotFlip18}, but this would    
increase the difficulty for realizing \adlof{} as a black-box attacker because different types of hot words may be relevant for different tasks. 

\subsection{Character Embeddings}
We consider three different character embedding spaces. 
The first is continuous, assigning each character a dense 576 dimensional representation, 
which allows, e.g., for computing cosine similarities between any two characters as well as nearest neighbors for each input character. 
The other two are discrete and merely used as arguments to \adlof{}. 
Thus, they are only required to specify nearest neighbors for standard input characters. For them, each character $c$ in a selected range 
(e.g., standard English alphabet a-zA-Z)
is assigned a set of nearest neighbors, and all nearest neighbors are equidistant to $c$. 
All three CES carry visual information, i.e., nearest neighbors are visually similar to the character in question. For practical reasons, we limit all our perturbations to the first 30k Unicode characters throughout. 

\paragraph{Image-based character embedding space (ICES)} provides a continuous image-based character embedding (ICE) for each Unicode character.
We retrieve a 24$\times$24 image representation of the character
(using Python's PIL library), then 
stack the rows of this matrix (with entries between $0$ and $255$) to form a $24\cdot 24 = 576$ dimensional embedding vector.

\paragraph{Description-based character embedding space (DCES)} is based on the textual descriptions of Unicode characters. 
We first obtain descriptions of each character from the Unicode 11.0.0 final names list (e.g., \textsc{latin small letter a} for the character `a').
Then we determine a set of nearest neighbors 
by choosing all characters whose descriptions refer
to the same letter in the same case, 
e.g., an alternative to \textsc{latin small letter a}
is \textsc{latin small letter a with grave} as it contains the 
keywords \textsc{small} and \textsc{a}.

\paragraph{Easy character embedding space (ECES)} provides manually selected simple visual perturbations.
It contains exactly one nearest neighbor for each of the 52 characters a-zA-Z, 
chosen as a diacritic below or above a character, such as $\hat{c}$ for 
the
character $c$.

\paragraph{Differences between the CESs}
The three embedding spaces play different roles in our experiments.
We use ICES as character representations in deep learning systems. DCES and ECES are used as input to \adlof{} to perturb our test data.\footnote{We do not attack with ICES because we also shield with ICES and this would be a (very unrealistic) white box defense scenario. Besides, ICES is also more difficult to restore for humans (see below), making it less desirable for an attacker.} ECES models a `minimal perturbance with maximal impact' scenario: 
we assume that ECES perturbations do not or only minimally affect human perception but may still have a large impact upon NLP systems. Indeed, we could have chosen an even simpler embedding space, e.g., by considering  visually identical characters in different alphabets, such as the Cyrillic `a' 
(Unicode 1072) for a Latin 
`a' 
(Unicode 97). 
DCES is a more difficult test-bed designed for evaluating 
our approaches under more realistic conditions with more varied and stronger attacks.  

Table \ref{table:ces} exemplifies the differences between ICES, DCES, and ECES by comparing the nearest neighbors of a given character. 
As expected, ICES contains neighbors of
characters which are merely visually similar without representing the
same underlying character (such as $\Lambda$ as a neighbor of A, or 
\includegraphics[scale=0.8]{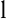}
as a neighbor of i). In contrast, DCES sometimes has neighbors with
considerable visual dissimilarity to the original character such as
Cyrillic small letter i 
(\includegraphics[scale=0.8]{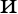})
which rather resembles a
mirror-inverted n. 
The overlap between ICES and DCES is
modest: out of 20 neighbors, a character has on average only four to five common neighbors in ICES and DCES.   

\begin{table*}
  \centering
  \includegraphics[scale=0.8]{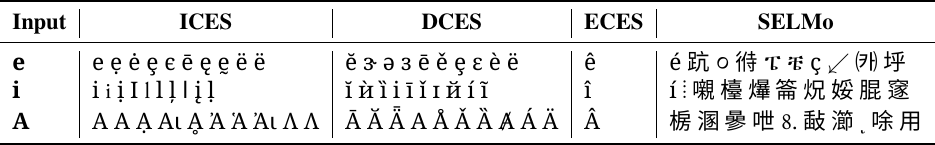}
  \caption{Ten nearest neighbors in our different character spaces. 
  `SELMo' refers to the nearest neighbors of the trained character embeddings in SELMo.}
  \label{table:ces}
\end{table*}

\subsection{Word Embeddings}
{Most neural NLP architectures encode text either on a
  character or word level.}
  For the latter, word embeddings are needed.
In this work, we use the ELMo architecture \cite{Peters:2018} to obtain (contextualized) word embeddings based on characters, i.e., there exists no fixed vocabulary and there will be no (word-level) out-of-vocabulary issues due to perturbation.
In the following, we outline our ELMo variant and a visual extension that includes visual signals from the input characters.
\paragraph{SELMo:}
ELMo as proposed by \citet{Peters:2018} first retrieves embeddings for every character in the input, which are learned as part of the network. ELMo then infers non-contextualized word embeddings by applying CNNs over all character embeddings in a word. Two layers of a deep bidirectional language model 
further process the word embeddings in their local sentential context and output contextualized word embeddings.

We slightly extend ELMo to include
character embeddings for the first 30k Unicode characters (instead of the default 256). 
We call this variant SELMo (``Standard ELMo'').
It is worth pointing out that the learned character embeddings of SELMo carry almost no visual information, as illustrated in Table~\ref{table:ces}. That is, except for a few very standard cases, nearest neighbors of characters do not visually resemble the orginal characters, even when trained on the 1 billion word benchmark {\cite{Chelba:2013}}.\footnote{We believe SELMo nearest neighbors are more likely to be Chinese/Japanese/Korean (CJK) characters because these nearest neighbors are largely random and there are far more CJK characters in our subset of Unicode.}  

\paragraph{VELMo:}
To obtain a visually informed variant of ELMo, we replace learned character
embeddings with the ICEs 
and keep the character embeddings fixed during training. This means that during training, the ELMo model learns to utilize visual features of the input, thus potentially being more robust against visual attacks.
We call this variant VELMo (``Visually-informed ELMo''). 

To keep training times of SELMo and VELMo feasible, we use an output dimensionality of 512 instead of the original ELMo's 1024d output. Our detailed hyperparameter setup is given in \S\ref{ssec:hyperparams}.

\section{Human annotation experiment}

\begin{table*}[thb]
\centering
\includegraphics[scale=0.8]{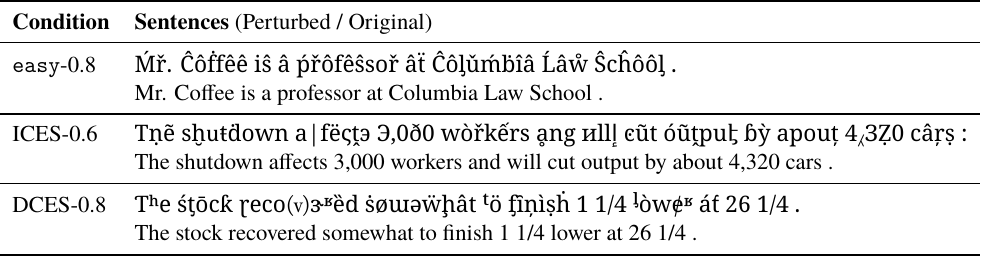}
\caption{Examples of perturbed sentences and underlying originals.}
\label{table:human}
\end{table*}

We asked 6 human annotators, 
{university employees and students} with native or near-native English language skills, to recover the original underlying English sentences given some perturbed 
text (data taken from the POS tagging and Chunking tasks, see Table \ref{table:tasks}). We considered different conditions:
\begin{itemize}[noitemsep]
  \item[(i)] \texttt{clean}: $\adlof{}(0,\_)$, i.e., no perturbation; 
    \item[(ii)] $\adlof{}(p,\text{ICES})$ for $p=0.2,0.4,0.6,0.8$;
   \item[(iii)] $\adlof{}(p,\text{DCES})$ for $p=0.2,0.4,0.6,0.8$;
   \item[(iv)] \texttt{easy}: $\adlof{}(p,\text{ECES})$ for $p=0.4,0.8$.
\end{itemize}\color{black}
{For each condition, we used 60-120 sentences, where at most 20 sentences of one condition were given to an annotator.}
Examples of selected conditions are shown in Table \ref{table:human}.
Our rationale for including this 
recovery task is to test robustness of human perception under (our) visual perturbations. We focus on recovery
instead of an extrinsic task such as POS because 
the latter would have required expert/trained annotators. 

We evaluate
by measuring the 
normalized edit distance between the recovered sentence and the underlying original, averaged over all sequence pairs and all human annotators. We normalize by the maximum lengths of the two sequences. In our case, this metric can be interpreted as the fraction of characters that have been, on average, wrongly recovered by human annotators. We refer to the metric as ``error rate''.

\begin{figure}[!htb]
\centering
\includegraphics[width=0.5\textwidth]{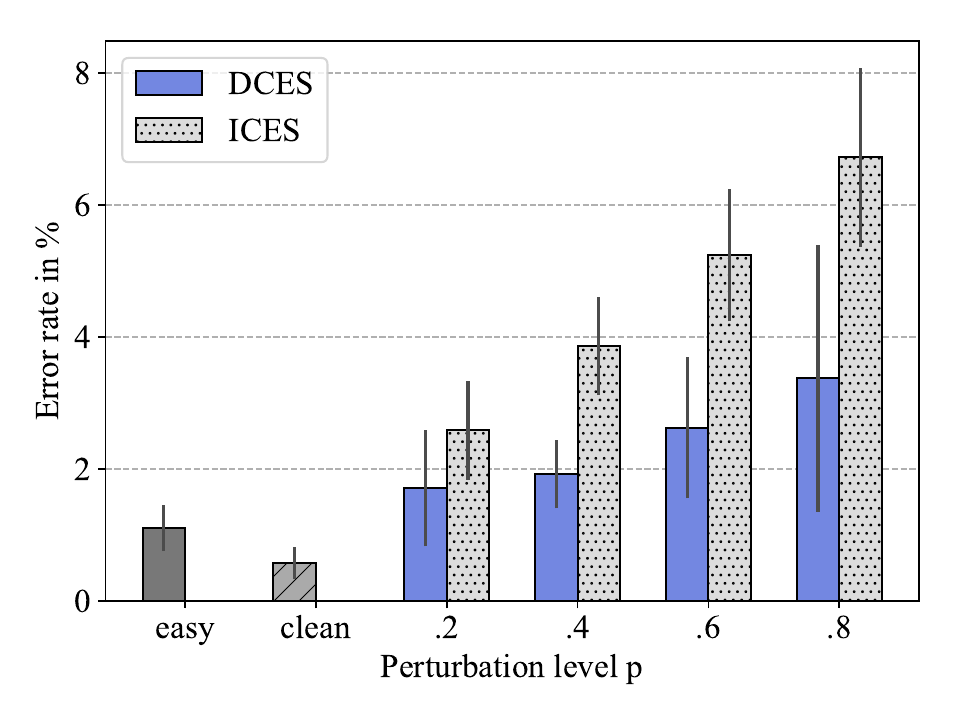}
\caption{Human annotation experiment. Error bars indicate 
std.\ 
across annotators. For \texttt{easy}, we merge the cases $p=0.4/0.8$.}
\label{fig:human}
\end{figure}

Results are shown in Figure \ref{fig:human}.
In \texttt{easy}, there is 
almost no difference between perturbation levels $p=0.4$ and $p=0.8$, so we merge the two conditions. 

Humans make copy mistakes even when the input is not perturbed, 
as evidenced by a positive error rate in \texttt{clean}. Such mistakes are typically misspellings or the wrong type of quotation marks (''~vs.~``).
We observe 
a slightly higher error rate in \texttt{easy} 
than in \texttt{clean}.
However,  
on average 75\% of all sentences are (exactly) correctly recovered in \texttt{easy} while this number is lower (72.5\%) in \texttt{clean}.  
By chance, \texttt{clean} contains fewer sentences with quotation marks than \texttt{easy}, for which a copy mistake was more likely. This may explain \texttt{easy}'s higher error rate. 

As we increase the perturbation level, the error rate increases consistently for DCES/ICES. 
It is noteworthy that DCES perturbations are easier to parse for humans than ICES perturbations. 
{We think} this is because 
DCES perturbations always retain a variant of the same character, while ICES may also disturb one character to another character (such as $h$ to $b$). 
Another explanation is that ICES, unlike DCES and ECES, also disturbs numbers and punctuation. 
Numbers, especially, are more difficult to recover.
However, even at 80\% disturbance level, humans can, on average, correctly recover at least 93\% of all characters in the input text in all conditions. 

In summary, 
humans appear very good at 
understanding visual perturbations, and are almost perfectly robust to the easy perturbations of ECES. 
Since adversarial attacks should have minimal impact on humans \cite{Szegedy:2014}, 
the good performance of humans especially on ECES and DCES makes these two spaces ideal candidates for attacks on NLP systems.

\section{Computational Experiments}
We now evaluate the capabilities of SOTA neural network models to deal with visual attacks in four extrinsic evaluation tasks described in \S\ref{sec:tasks} and illustrated in Table~\ref{table:tasks}. Hyperparameters of all our models are given in \S\ref{ssec:task_settings}. 
We first examine the robustness of all architectures to visual perturbations 
in \S\ref{sec:robustness}
and then evaluate different shielding approaches in \S\ref{sec:adv}.

\subsection{Tasks}\label{sec:tasks} 
\label{sec:experiments:tasks}
\begin{table*}[!htb]
\centering
\includegraphics[scale=0.8]{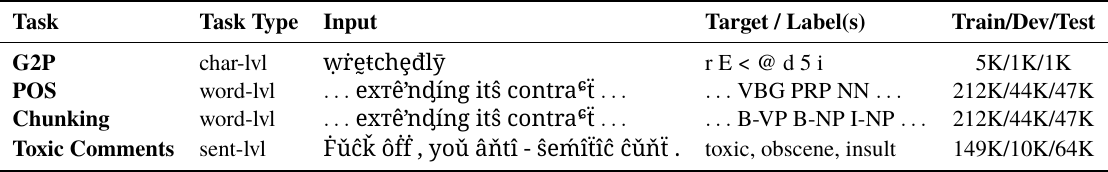}
\caption{NLP tasks considered in this work, along with (perturbed) examples and data split statistics. }
\label{table:tasks}
\end{table*}
\paragraph{G2P:}
As our first task, we consider the character-level task %
of grapheme-to-phoneme (G2P) conversion. It consists of %
transcribing a character input stream into a phonetic representation. 
As our dataset, we choose the Combilex pronunciation dataset of American English
\cite{Richmond:2009}.
We frame G2P as a sequence tagging task. To do so, we first hard-align input and output sequences using a 1-0,1-1,1-2 alignment scheme \cite{Schnober:2016} in which an input character is matched with zero, one, or two output characters. 
Once this preprocessing is done, input and output sequences have equal lengths and we can apply a standard BiLSTM on character-level to the aligned sequences \cite{Reimers:2017}.

\paragraph{POS \& Chunking:} 
We consider two word-level tasks.
POS tagging associates each token with its corresponding word class 
(e.g., \textit{noun, adjective, verb}). Chunking 
groups words into syntactic chunks such as noun and verb phrases (NP and VP), assigning a unique tag to each word, which encodes the position and type of the syntactic constituent, e.g., begin-noun-phrase (B-NP). 
We use the training, dev and test splits provided by the CoNLL-2000 shared task  \cite{SangB00}
and
use 
the same BiLSTM architecture as above with SELMo/VELMo embeddings. 

\paragraph{Toxic comment (TC) classification:} 
A very realistic use case for adversarial attacks is the toxic comment classification task. 
One could easily think of a scenario where a person with malicious intent explicitly aims to fool automated methods for detecting toxic comments or insults by obfuscating text with non-standard characters that are still human-readable. 
We conduct experiments on the TC classification task provided by Kaggle.\footnote{https://www.kaggle.com/c/jigsaw-toxic-comment-classification-challenge/} 
It is a multi-label sentence classification task 
with six classes, i.e., \textit{toxic}, \textit{severe toxic}, \textit{obscene}, \textit{threat}, \textit{insult}, \textit{identity hate}.
We use average SELMo/VELMo embeddings as input to an MLP.

\begin{figure}[!htb]
\centering
  \includegraphics[scale=0.85]{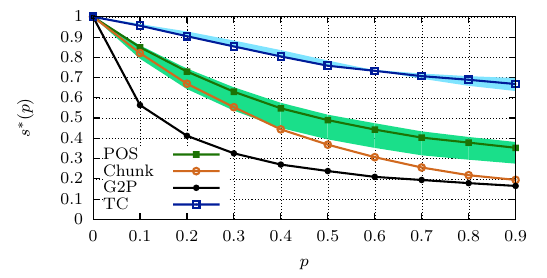}
  \caption{Degradation of SOTA systems for different perturbation levels when attacked by \adlof{}($p$,DCES). The colored regions show how the performance of other SOTA systems relate to ours (i.e., they all suffer from similar degradation).}
  \label{fig:sota}
\end{figure}

\subsection{\adlof{} attacks }\label{sec:robustness}
In Figure \ref{fig:sota}, we plot how various SOTA systems degrade as we perturb the test data using DCES. 
We do not only include our own systems, but also existing SOTA models: Marmot \cite{Mueller:2013} and Stanford POS tagger (SPT)~\cite{coreNLP:14}. Marmot is a feature-based POS tagger and trained on our data splits. SPT is a bi-directional dependency network tagger that mostly employs lexical features. For SPT, we used the pretrained English model provided by the toolkit. 
Further, we include a FastText TC classifier which has achieved SOTA performance.\footnote{https://www.kaggle.com/yekenot/pooled-gru-fasttext} We additionally experiment with word level dependency embeddings for POS tagging and TC classification \cite{Komninos:2016}. 

To compare the performance of different tasks, Figure \ref{fig:sota} shows
scores computed by:
\begin{align*}
 s^{*}(p) = \frac{s(p)}{s(0)}, 
\end{align*}
where $p$ is the perturbation level and $s(p)$ is the score for each task at $p$, measured in 
edit distance for G2P, accuracy for POS tagging, micro-F1 for chunking, and AUCROC for TC classification. We invert the scores $g$ of G2P by $1/g$ since lower scores are better for edit distance. 
Thus, 
$s^*(0)$ is always 1 and $s^*(p)$ is 
the relative performance compared to the clean case of no perturbations. 

We see that all systems degrade considerably. For example, all three POS taggers have a performance of below 60\% of the clean score when 40\% of the input characters are disturbed. 
Chunking degrades even more strongly, and 
G2P has the highest drop: 10\% perturbation level causes a 40\% performance deterioration.
This may be because G2P is a character-level task and the perturbation of a single character is analogous to perturbing a complete word in the word-level tasks. Finally, TC classification degrades least, i.e., only at $p=0.9$ do we see a degradation of 30\% relative to the clean score. These results appear to suggest that character-level tasks suffer the most from our \adlof{} attacks and sentence-level tasks the least. However, it is worthwhile pointing out that lower-bounds for individual tasks may depend on the evaluation metric (e.g., AUCROC always yields 0.5 for majority class voting) as well as task-specific idiosyncrasies such as the size of the label space. 

We note that the degradation curves look virtually identical for both DCES or ECES perturbations (given in \S\ref{ssec:eces_results}). %
This is in stark contrast to human performance, where ECES was much easier to parse than DCES, indicating the discrepancies between human and machine text processing. 

\subsection{Shielding}\label{sec:adv}
We study 
four forms of shielding against \adlof{} attacks: adversarial training (\textbf{AT}), {visual} 
character embeddings (\textbf{CE}), \textbf{AT$+$CE}, and rule-based recovery (\textbf{RBR}).
For \textbf{AT}, we include visually perturbed data at train time. 
We do not augment the training data, but 
replace clean examples using \adlof{} in the same way as for the test data. Based on preliminary experiments with the G2P task, we apply \adlof{} to the training data using $p_{\text{train}}=0.2$. Higher levels of $p_{\text{train}}$ did not appear to improve performance. For \textbf{CE},
we 
use fixed ICEs, either fed directly into a model (G2P) or via VELMo (all other tasks). For \textbf{AT}$+$\textbf{CE}, we combine adversarial training with visual embeddings. Finally, for \textbf{RBR}, we replace each non-standard character in the input stream with 
its nearest standard neighbor in ICES, where we define the standard character set as 
a-zA-Z plus punctuation.   

Rather than absolute scores, we report 
differences between the scores in one of the shielding treatments 
and original scores:
\begin{align*}
  \Delta_{\tau}:=\sigma^*(p)-s^*(p),\quad \sigma^*(p):=\sigma(p)/s(0) 
\end{align*}
where $\sigma(p)$ is the score for each task using a form of shielding. The value 
$\Delta_{\tau}$ denotes the improvement of the 
scores from shielding method $\tau$ over the original scores without shielding. We normalize $\sigma(p)$ by the score $s(0)$ of the systems without shielding on clean data. We also note that our 
test perturbations 
are unseen during training 
for DCES; for ECES this would not make sense, because each character has only one nearest neighbor. In the following, we report results mostly for DCES and show the ECES results in \S\ref{ssec:eces_results}. We highlight marked differences between the results, however. 
\begin{figure*}
\centering
    \includegraphics{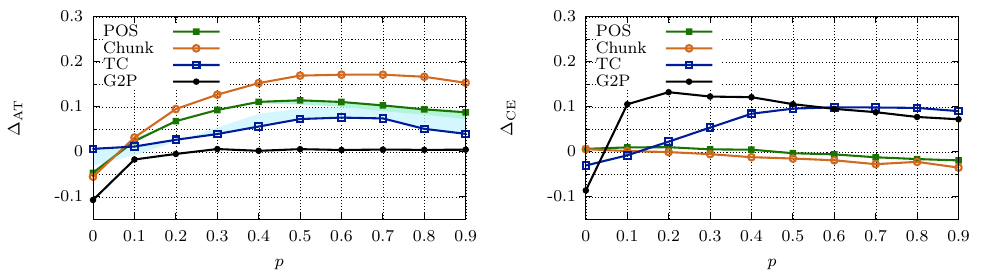}
  \caption{AT (with ICES replacements) and CE tested on DCES perturbed data. The colored regions show AT (with random replacements).}
  \label{fig:adversarial}
\end{figure*}

All tasks typically profit considerably from \textbf{AT} (Figure \ref{fig:adversarial} left). Chunking scores improve most; e.g., at $p=0.5$,  $\sigma^*$ is 
{17 percent points (pp)} higher than $s^*$. 
AT does not help for G2P in the DCES setting but it does help for ECES (see \S\ref{ssec:eces_results}), where test perturbations may have been seen during training. 
We 
conjecture 
that AT makes systems generally aware that the input can be broken in some way and forces them to shield against such situations, an effect similar to dropout. However, such shielding appears more difficult in character-level tasks, where a missing token is considerably more damaging than in word-  or sentence-level tasks.

In Figure \ref{fig:adversarial} (right), we observe that \textbf{CE} helps a lot for G2P, but much
less 
particularly for POS and Chunking. We believe 
that for G2P, the visual character embeddings restore part of the input and thus have considerable effect. It is surprising, however, that visual embeddings have no positive effect for both word-level tasks, and instead lead to small deteriorations. A possible explanation is that, as the character embeddings are fed into the ELMo architecture, their effect is dampened. Indeed, 
we performed a sanity check (see \S\ref{ssec:intrinsic-evaluation}) to test how (cosine) similar a word or sentence $w$ is to a perturbed version $w'$ of $w$ under both SELMo and VELMo. We found that VELMo assigns consistently better similarities but the overall gap is small. 

\begin{figure*}
\centering
    \includegraphics{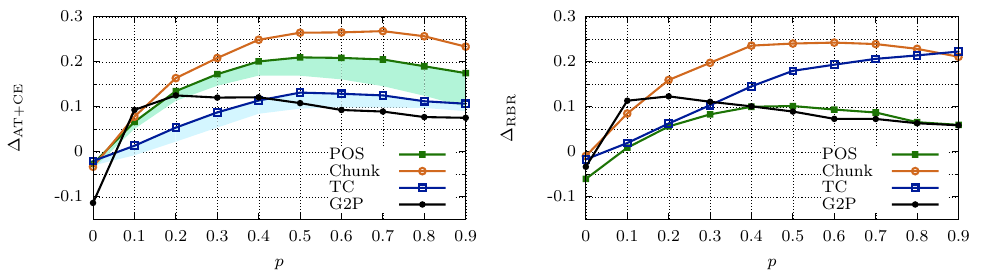}
\caption{AT$+$CE (with ICES replacements) 
and RBR on DCES perturbed data. The colored regions show AT (with random replacements).} 
  \label{fig:combined_plain}
\end{figure*}

We observe that the combined effect of AT and CE  (\textbf{AT}$+$\textbf{CE}, Figure~\ref{fig:combined_plain} left) is always substantially better than either of the two alone. For instance, at $p = 0.5$, POS improves by about
20pp, while AT alone had
an effect of only 12pp and the effect of 
CE was even negative. Thus, it appears that AT is able to kick-start the benefits of CE, especially in the case when they alone are not effective.

\textbf{RBR} is excellent for ECES (see \S\ref{ssec:eces_results}). It has a small negative effect on clean data, meaning that there is some foreign material in English texts which gets corrupted by RBR, but 
for any $p>0$ the performance under RBR is almost on the level of $p=0$ for ECES. 
RBR is also consistently better than CE, even though both depend on ICES: CE in a `soft' way and RBR in a `hard' way. Our best explanation is that RBR is analogous to `machine translating' a foreign text into English and then applying a trained classifier, while CE is analogous to a direct transfer approach \cite{McDonald:2011} which trains in one domain and is then applied to another. This causes a form of domain shift to which neural nets are quite vulnerable 
\cite{Ruder:2018,Eger:2018:Coling}. 

For DCES, RBR is 
outperformed by AT$+$CE, which better mitigates the domain shift than CE, except for TC. 

We note that even with all our shielding approaches, the performance of the shielded systems is still considerably below the performance on clean data at some perturbation levels. E.g., at $p=0.9$, AT$+$CE shielded Chunking has a score of less than 60\% of the clean performance. While it may be partially due to our character embeddings not being optimal (i.e., they assign low similarity to major and minor variants of the same letter such as 
\includegraphics[scale=0.8]{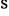}
and s,  
which could be improved by size-invariant CNNs), a main reason for this could be the domain-shift induced by the perturbations, for which even AT cannot always help when attacks are unseen during training. This is another major distinction between human and machine processing. 

\section{Discussion}\label{sec:discussion}
\begin{table*}[!htb]
	\centering
        \includegraphics[scale=0.8]{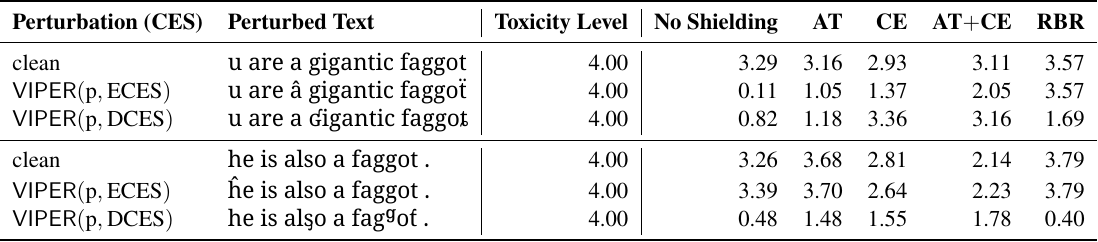}
        \caption{Two examples of toxic/non-toxic comments that show the effects of the different shielding methods. We report the averaged sum over the six toxicity classes, e.g., {$4.0$} is equal to a positive example in four classes. $p=0.1$.}
        \label{tab:toxic-comments-dces-pos}
\end{table*}

\paragraph{Random AT.} 
We discussed that the effect of AT may be similar to 
dropout. 
If so, AT with random rather than visual character replacements 
should be similarly effective. 
Indeed, the graphs in Figures \ref{fig:adversarial} and \ref{fig:combined_plain} show 
that injecting random noise generally improves the robustness, but not to the level of visually informed AT (exemplarily for POS/TC). 
\paragraph{Error analysis.}
We analyze the cases in which our perturbation with \adlof{}  changes the prediction to a wrong class. We perform our analysis for TC as it 
represents a real-world attack scenario.
We define the \textit{toxicity level} (TL) of a comment~$x$ 
for a class 
$y\in\{1,\ldots,6\}$
and model 
$\theta$ as 
$\text{TL}(x) = 
\sum_y \theta(x,y),$ 
e.g., a comment $x$ which has been classified as \textit{insult} (probability 0.8)
and \textit{obscene} (probability 0.7) gets a TL of $1.5$. 
We consider a successful attack to decrease TL after perturbation. 
\adlof{} with DCES and ${p=0.1}$ achieves a success rate of 24.1\%---i.e., roughly one fourth of the \textit{toxic} comments receive a lower TL. 
In contrast, the impact on \textit{non-toxic} comments is small---
TL increased in only 3.2\% of the cases.

Table~\ref{tab:toxic-comments-dces-pos} shows sample comments and their TL for different shielding and perturbation methods. As can be seen, perturbing specific words (\emph{hot words} for TC)  substantially reduces the TL score of a non-shielded approach (e.g., from 3.29 to 0.11), while perturbing `non-hot' words like `he' has little effect. The shielding approaches help in these show-cased examples to various degrees and the shielding with AT$+$CE is more robust to stronger attacks (higher visual dissimilarity) than RBR. 

This illustrates that a malicious attacker may aim to increase the success rate of an attack by only perturbing offensive words (in the TC task).
To test whether \adlof{} benefits 
from perturbing such hot words, we manually compiled a list of 20 hand-selected offensive words 
(see \S\ref{ssec:wordlist}) which we believe are indicators of toxic comments. We then analyzed how often a perturbation of a word from this list co-occurs with a successful attack. 
We observe that in 55\% of successful attacks, 
a word from our list was among the perturbed words of the comment.
As our list is only a small subset of all possible offensive words, 
the perturbation of hot words may have an even stronger effect.

\section{Conclusion}\label{sec:conclusion}
In this work, we considered visual modifications to text as a new type of adversarial attack in NLP and we showed that humans are able to reliably recover visually perturbed text.
In a number of experiments on character-, word-, and sentence-level, we highlighted the fundamental differences between humans and state-of-the-art NLP systems, which sometimes blatantly fail under visual attack, showing that visual adversarial attacks can have maximum impact.
This calls for models that have richer biases than current paradigm types do, which would allow them to bridge the gaps in information processing between humans and machines. 
We have explored one such bias, visual encoding, 
but our results suggest that further work on such shielding is necessary in the future.

Our work is also important for system builders, 
such as 
of 
toxic comment  detection 
models deployed by, e.g., Facebook and Twitter,  
who regularly face visual attacks, 
and who might face even more such attacks once visual character perturbations are easier to insert than via the keyboard. 
From the %
opposite 
viewpoint, \adlof{} may 
help users retain privacy in online engagements and when trying to avoid censorship   \cite{Hiruncharoenvate:2015} by suggesting visually similar spellings of words. 

Finally, our work shows that the `brittleness'  \cite{Belinkov:2018} of NLP extends beyond MT and beyond word reordering or replacements, a   
 recognition that 
 we hope inspires others to investigate more ubiquitous  shielding techniques.

\subsubsection*{Acknowledgments}
We thank the reviewers for helpful feedback. 
This work has been supported by the German Research Foundation (DFG) funded research training group ``Adaptive Preparation of Information form Heterogeneous Sources'' (AIPHES, GRK 1994/1), the DFG-funded projects QA-EduInf (GU 798/18-1, RI 803/12-1), DIP (GU 798/17-1), the German Federal Ministry of Education and Research (BMBF) under the promotional references 16DHL1040 (FAMULUS) and by the Hessian research excellence program “Landes-Offensive zur Entwicklung Wissenschaftlich- \"Okonomischer Exzellenz” (LOEWE) as part of the a! - automated language instruction project (No. 521/17-03). 
We gratefully acknowledge support of NVIDIA Corp.\ with the donation of the Tesla K40 GPU used for this research.
Calculations for this research were also conducted on the Lichtenberg high performance cluster of Technical University Darmstadt. 
%

\bibliography{naacl2019}
\bibliographystyle{acl_natbib}

\clearpage
\appendix

\section{Appendices}
\label{sec:appendix}

\subsection{SELMo and VELMo Hyperparameters}
\label{ssec:hyperparams}
Differences to the original ELMo as in \cite{Peters:2018} are:
\begin{itemize}[noitemsep]
\item We exclude CNN filters of size 6 and 7.
\item The maximum characters per token is 20 (instead of 50).
\item The LSTM dimensionality is 2048 (instead of 4096).
\item Our projection dimensionality is 256 (instead of 512).
\item We train the models for 5 epochs (instead of training it for 10 epochs).
\end{itemize}

\subsection{Task Settings}
\label{ssec:task_settings} 
\paragraph{G2P:} 
We randomly draw most hyperparameters for the sequence tagging BiLSTM architecture \cite{Reimers:2017} that we use for G2P, e.g., those concerning the optimizer used, learning and dropout rates. We hand-set the number of hidden recurrent layers to 1, and its size to 50. We use early stopping and set the maximum number of epochs for training to 50. As our dataset, we choose the Combilex pronunciation dataset of American English \cite{Richmond:2009}. 
We randomly draw our train/dev/test splits from the whole corpus. Examples and split sizes are given in Table 
\ref{table:tasks}. 
We report \textbf{edit distance} between desired pronunciations and predicted pronunciations as metric. We report the edit distance averaged across all 1k test strings, averaged over 5 random initializations of all weight matrices.

\begin{figure}[!htb]
\centering
    \includegraphics[scale=0.9]{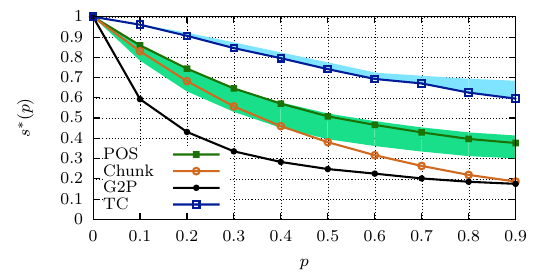}
  \caption{Degradation of SOTA systems for different perturbation levels when attacked by \adlof{}($p$,ECES). The colored regions show how the performance of other SOTA systems relate to ours.}
  \label{fig:eces-sota}
\end{figure}

\paragraph{POS \& Chunking:} 
We use the training, dev and test splits provided by the CoNLL-2000 shared task~\cite{SangB00} for both tasks. We have used a readily available LSTM-CRF sequence tagger~\cite{Reimers:2017} as above, but adapted for ELMo-type input embeddings, with default hyperparameter settings. We run each experimental setting 10 times, and report the average, measured as \textbf{accuracy} for POS and \textbf{micro-F1} for Chunking. For both tasks, we have used two stacked BiLSTM-layers with 100 recurrent units and dropout probability of $0.5$. Mini-batch size is chosen as 32. We used gradient clipping and early stopping to prevent overfitting. Adam is used as the optimizer.

\paragraph{Toxic comment classification:} 
We use the train and test splits provided by the task organizers. 
For tuning our models, we split off a development set of 10k sentences from the training data. 
As in POS\&Chunking, we train models on clean and perturbed data using SELMo and VELMo representations. 
We obtain the sentence representation for a single sentence by averaging ELMo word embeddings over all tokens. 
We then train an MLP which we tune separately for each SELMo and VELMo embedding using random grid search 
with 100 different configurations. 
We tune the following hyperparameters separately for each hidden layer: the depth of the neural network, i.e., one, two, or three hidden layers; the size of the hidden layer (128, 256, 512, or 1024); the amount of dropout after each layer (0.1 - 0.5); the activation functions for each hidden layer (\textit{tanh}, \textit{sigmoid}, or \textit{relu}) \cite{Eger:2018:EMNLP}.
Both models are trained for 100 epochs with an early stopping after 10 epochs without any substantial improvement and use Nesterov-accelerated Adaptive Moment Estimation \cite{Dozat2016} for optimization.
Model performance is measured as proposed by the task organizers using the \textit{area under the receiver operating characteristics curve} (\textbf{AUCROC}). 

\subsection{ECES Results}
\label{ssec:eces_results}

\begin{figure*}
\centering
    \includegraphics{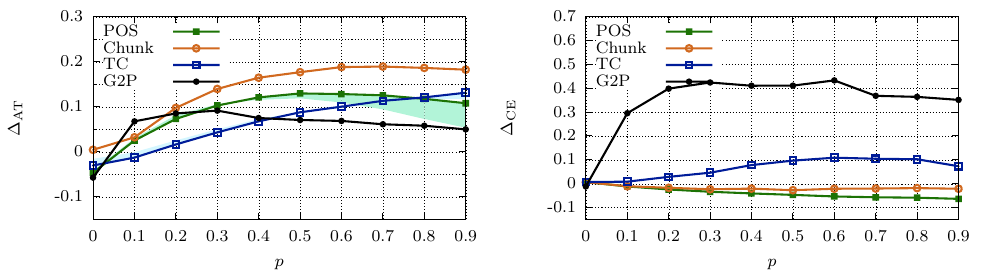}
  \caption{AT (ICES) and CE tested on ECES perturbed data. The colored regions show AT (Random). The y axis of $\Delta_\mathrm{AT}$ spans $-0.15$-$0.3$ for better visualization.}
  \label{fig:adversarial-eces}
\end{figure*}

\begin{figure*}
\centering
    \includegraphics{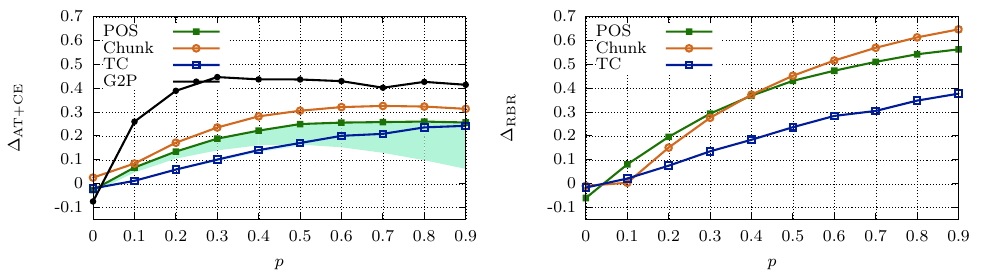}
\caption{AT$+$CE (ICES) and RBR on ECES perturbed data. The colored regions show AT (Random).} 
  \label{fig:ad-rbr-eces}
\end{figure*}

Figure~\ref{fig:eces-sota} shows how the performances of various SOTA systems degrade on ECES settings.
Figures~\ref{fig:adversarial-eces} and \ref{fig:ad-rbr-eces} show our shielding results on ECES perturbed data. 
As indicated in the main paper, RBR is able to recover ECES data almost perfectly regardless of the perturbation level. 
This is because ECES only perturbs with a single nearest neighbor, which in addition is visually extremely similar to the underlying original, and thus, RBR can almost completely undo the perturbations. 

\subsection{AT$+$CE \textit{vs.} AT or CE}
\label{ssec:atce_vs_atorce}
\begin{figure*}
  \centering
    \includegraphics{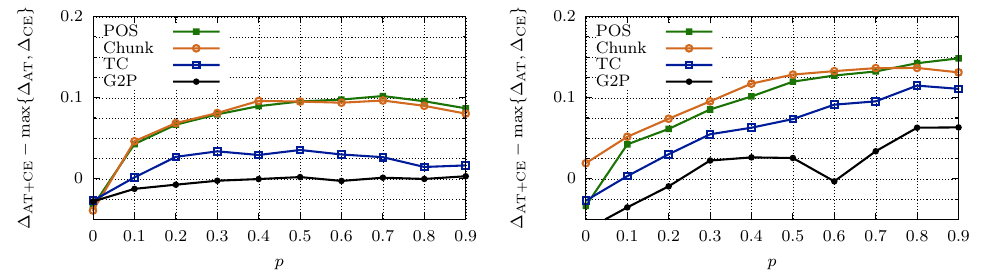}
  \caption{AT$+$CE (ICES) vs max(AT,CE) for DCES and ECES perturbed test data.}
  \label{fig:visad_vis-ad}
\end{figure*}

Figure~\ref{fig:visad_vis-ad} compares AT$+$CE against either AT or CE. For this, we compute the difference in the performance decrease normalized by the test performance on the clean data. 
As can be seen, AT$+$CE almost constantly outperforms either one of both, especially on word- and sentence-level tasks. 

\subsection{Intrinsic Evaluation}
\label{ssec:intrinsic-evaluation}

\begin{figure*}[!htb]
\centering
    \includegraphics{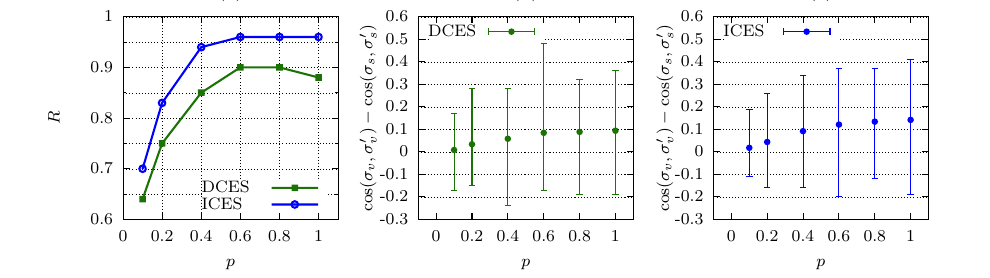}
\caption{Results of the intrinsic evaluation. (a) shows the ratio of cases in which the condition in Eq.~(\ref{eq:cos_vs}) is met (`$R$'). (b) and (c) show the average difference of the cosine similarities when sentences are perturbed with (b) DCES and (c) ICES.} 
  \label{fig:sanity_check}
\end{figure*}

To analyze the differences between VELMo and SELMo, we investigate whether the models learn similar word embeddings for a clean sentence and its visually perturbed counterpart. We compare sentence embeddings which we obtain by averaging over the SELMo or VELMo word embeddings of a sentence (clean or perturbed). 

\paragraph{Setup}
Given a sentence embedding $\sigma$ of a clean sentence and an embedding $\sigma^\prime$ of its visually perturbed counterpart, obtained by either averaging over VELMo (indicated by subscript $v$) or SELMo (indicated by subscript $s$) word embeddings, we test if the condition
\begin{equation}
	\label{eq:cos_vs}
  \cos(\sigma_v,\sigma_v^\prime)>\cos(\sigma_s,\sigma_s^\prime)
\end{equation}
is met (with $\cos$ being the cosine similarity). 

For our experiments, we randomly sample 1000 sentences from the TC dataset (see \S\ref{sec:experiments:tasks}) 
and perturb them with \adlof{}($p$, CES) where CES  $\in\lbrace$ICES, DCES$\rbrace$. 
We then count the number $N$ of cases in which the above condition is met with regards to the chosen CES and the value of $p$, and report the ratio $R=N/1000$.

\paragraph{Results}  
The results are given in Figure~\ref{fig:sanity_check}. In Figure~\ref{fig:sanity_check}(a) we observe that the VELMo embeddings of a clean sentence and its perturbed counterpart are in many cases more similar than the ones of SELMo. For larger values of $p$, this ratio substantially increases from 70\% to 95\% (with ICES), which shows that VELMo is better suited to capture the similarity to the source sentence, especially in cases with a strong perturbation. 

In Figures~\ref{fig:sanity_check}(b) and (c) we show the (mean) 
difference 
$\cos(\sigma_v, \sigma_v^\prime) - \cos(\sigma_s,\sigma_s^\prime)$.
Here, we only observe a small positive effect in favor of VELMo, showing that the advantage of VELMo over SELMo is not substantial. We hypothesize that this is due to the contextual information which is utilized throughout the ELMo architecture, allowing SELMo to infer individual characters from the context of the word and the sentence. However, our results also show that the advantage of VELMo over SELMo is consistent.

Differences in similarities can also be affected by 
model training or the model architecture---e.g., in an extreme case a model could output the same embedding for every word/sentence. 
This would result in a `perfect' cosine similarity, which would be  advantageous in the previous experiment. Thus, we perform an additional experiment where we examine if 
\begin{equation}
	\cos(\sigma_v,\sigma_v^\prime)>\cos(\sigma_v,\rho_v)
\end{equation}
holds, where $\rho$ is a randomly sampled sentence from the TC dataset (with no perturbation). The same experiment is also performed for SELMo.

The results in Table~\ref{table:sanity_check_random_experiment} show that for both models with $p=0.1$ the original sentence is in 97--100\% of the cases more similar to its perturbed counterpart than the randomly chosen sentence. As the probability of perturbed characters increases, VELMo has a clear advantage over SELMo. For example, if we perturb all characters in a sentence ($p=1.0$), the SELMo embeddings of the perturbed sentence are in 1\% of the cases more similar to the original sentence whereas this is the case for more than 34--39\% for VELMo.
Thus, VELMo embeddings better capture the similarity between visually similar words.

\begin{table}
	\small
    \begin{center}
        \begin{tabular}{@{}ccccc@{}}
            \toprule
                  & \multicolumn{2}{c}{\bf DCES} & \multicolumn{2}{c}{\bf ICES} \\ 
            \bf $p$       & \bf VELMo & \bf SELMo & \bf VELMo & \bf SELMo \\
            \midrule
            $0.1$ & $0.99$        & $0.98$          & $1.00$ & $0.97$         \\
            $0.2$ & $0.98$          & $0.84$        & $0.98$ & $0.77$            \\
            $0.4$ & $0.81$          & $ 0.38$        & $0.85$ & $0.25$         \\
             $0.6$ & $0.60$          & $ 0.10$        & $0.63$ & $0.07$         \\
            $0.8$ & $0.42$          & $0.04$          & $0.49$ & $0.03$            \\
            $1.0$ & $0.34$          & $0.01$         & $0.39$ & $0.01$            \\
            \bottomrule
        \end{tabular}
        \caption{Results of the intrinsic evaluation where we compare clean sentences to their perturbed counterparts as well as randomly chosen sentences. The numbers show the ratio of cases where clean sentences are more similar to their perturbed counterparts than the randomly chosen sentences.
        } 
        \label{table:sanity_check_random_experiment}
    \end{center}
\end{table}

\subsection{List of hand-selected curse words}
\label{ssec:wordlist}
{ arrogant, ass, bastard, bitch, dick, die, fag, fat, fuck, gay, hate, idiot, jerk, kill, nigg*\footnote{Due to several variations in the data, we match against \textit{nigg*} instead of the whole word.
}, shit, stupid, suck, troll, ugly}

\end{document}